\documentclass{vgtc}                          

\ifpdf
  \pdfoutput=1\relax                   
  \usepackage{graphicx}                
  \DeclareGraphicsExtensions{.pdf,.png,.jpg,.jpeg} 
\else
  \usepackage{graphicx}                
  \DeclareGraphicsExtensions{.eps}     
\fi%

\graphicspath{{figures/}{pictures/}{images/}{./}} 

\usepackage{microtype}                 
\PassOptionsToPackage{warn}{textcomp}  
\usepackage{textcomp}                  
\usepackage{mathptmx}                  
\usepackage{times}                     
\usepackage{cite}                      
\usepackage{tabu}                      
\usepackage{booktabs}                  
\usepackage{amsmath}
\usepackage{multirow}

\usepackage{floatrow}

\graphicspath{{./figures/}}
\usepackage{enumitem}
\usepackage{todonotes}

\usepackage{float}
\usepackage{dblfloatfix} 

\usepackage[utf8]{inputenc}

\newenvironment{s_itemize}{
\begin{itemize}[leftmargin=*]
  \setlength{\itemsep}{3pt}
  \setlength{\parskip}{0pt}
  \setlength{\parsep}{0pt}
}{\end{itemize}}

\newcommand{\changetext}[1]{\textcolor{black}{#1}}
\newcommand{\changetextt}[1]{\textcolor{black}{#1}}

\newenvironment{s_enumerate}{
\begin{enumerate}[wide, labelwidth=!, labelindent=0pt]
  \setlength{\itemsep}{2pt}
  \setlength{\parskip}{0pt}
  \setlength{\parsep}{0pt}
}{\end{enumerate}}
\onlineid{3154}

\vgtccategory{Research}

\vgtcinsertpkg

\title{Encode-Store-Retrieve: Augmenting Human Memory through Language-Encoded Egocentric Perception}

\author{Junxiao Shen\thanks{e-mail: js2283@cantab.ac.uk} %
\and John J. Dudley\thanks{e-mail: jjd50@cam.ac.uk} %
\and Per Ola Kristensson\thanks{e-mail: pok21@cam.ac.uk}}
\affiliation{\scriptsize University of Cambridge }

\teaser{
  \centering
  \includegraphics[width=\linewidth]{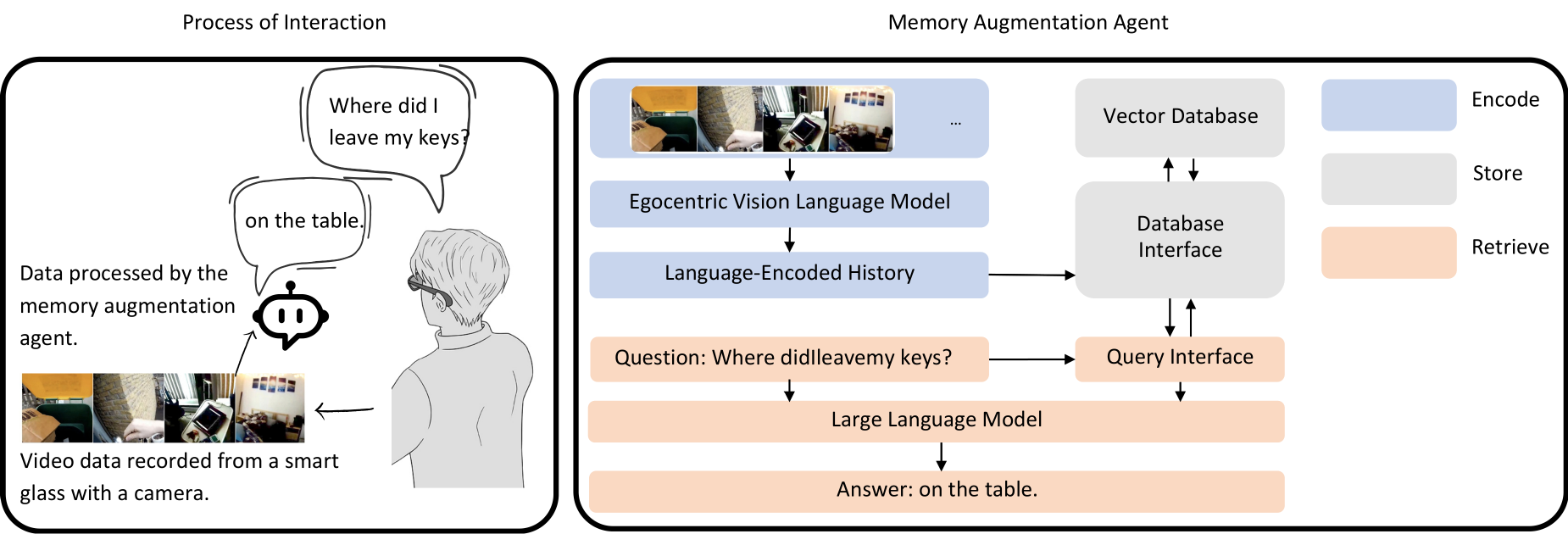}
  \caption{\textbf{Left}: Demonstration of a user interact with the memory augmentation agent. \textbf{Right}: The operation of the memory augmentation agent using language-encoded egocentric perception unfolds as follows. Initially, egocentric videos are encoded into linguistic representations using our bespoke large multimodal model, an egocentric vision language model. These language-encoded outputs are then held in a buffer, segmented into numerous chunks, featured as embeddings, and uploaded to a vector database. When a user poses a question---a common memory augmentation task such as ``Where did I leave my keys?''---the process begins by taking the user's query to generate a corresponding embedding. This produced embedding vector is then used to query the Database Interface with the intention of identifying relevant chunks related to the question by using vector similarity search algorithms. The retrieved chunks are subsequently amalgamated with the question to form a prompt for the large language model. The final response is then generated and presented to the user. 
  }
  \label{fig:teaser}
}

\abstract{We depend on our own memory to \textbf{encode}, \textbf{store}, and \textbf{retrieve} our experiences.
However, memory lapses can occur. 
One promising avenue for achieving memory augmentation is through the use of augmented reality head-mounted displays to capture and preserve egocentric videos, a practice commonly referred to as lifelogging.
However, a significant challenge arises from the sheer volume of video data generated through lifelogging, as the current technology lacks the capability to encode and store such large amounts of data efficiently. Further, retrieving specific information from extensive video archives requires substantial computational power, further complicating the task of quickly accessing desired content.
To address these challenges, we propose a memory augmentation agent that involves leveraging natural language encoding for video data and storing them in a vector database. This approach harnesses the power of large vision language models to perform the language encoding process. Additionally, we propose using large language models to facilitate natural language querying.
Our agent underwent extensive evaluation using the QA-Ego4D dataset and achieved state-of-the-art results with a BLEU score of 8.3, outperforming conventional machine learning models that scored between 3.4 and 5.8. 
Additionally, we conducted a user study in which participants interacted with the human memory augmentation agent through episodic memory and open-ended questions.
\changetextt{
The results of this study show that the agent results in significantly better recall performance on episodic memory tasks compared to human participants.
}
The results also highlight the agent's practical applicability and user acceptance.
} 

\CCScatlist{
  \CCScatTwelve{Human-centered computing}{Human computer interaction (HCI)}{Interaction paradigms}{Mixed / augmented reality};
  \CCScatTwelve{Information systems}{Information retrieval}{Retrieval tasks and goals}{Question answering};
    \CCScatTwelve{Computing methodologies}{Artificial intelligence}{Computer vision}{Computer vision tasks}{Visual content-based indexing and retrieval}
}

\begin{document}


\firstsection{Introduction}

\maketitle

\changetext{
Augmented reality (AR) enhances user perception by superimposing virtual content onto their environment, traditionally augmenting vision with digital overlays on real-world scenes~\cite{milgram1994taxonomy, azuma1997survey}. However, AR extends beyond visual augmentation to include auditory and tactile feedback~\cite{creed2024inclusive, liu2018augmented, lo2021navigation, katz2012navig, troncoso2020aiguide}. These multimodal capabilities make AR accessible and beneficial for a wider audience, enabling immersive and interactive experiences for all users, regardless of their abilities. This paper introduces a human memory augmentation agent that utilizes both the audio feedback and the egocentric front-facing camera of an AR device.
}

Human memory, crucial for various cognitive functions, often relies on external aids like photographs and reminders for enhancing recollection of past events~\cite{klein2015memory,intons1992external,intons2014external}. However, these aids are limited in scope. Some researchers suggest that lifelogging can be employed to augment memory~\cite{harvey2016remembering,dingler2021memory,chen2010augmenting}. Lifelogging involves the capture of images or videos, along with a wide array of personal data such as one's location, audio recordings, and physiological data~\cite{hayes2004personal,hodges2006sensecam,gurrin2014lifelogging}.

The introduction of smart glasses, including AR headsets with cameras, has enhanced lifelogging by enabling seamless capture of egocentric data along with additional sensor inputs. However, using these devices to enhance memory presents challenges, notably in efficiently encoding and storing the large, complex video data to maintain quality and enable real-time processing. Additionally, compliance with data privacy regulations is essential~\cite{gurrin2014privacy}.
Once the data has been encoded and stored, the next challenge is to ensure that this information can be retrieved in a useful and timely manner. 
Therefore, we need to design a system that is able to process large volumes of complex data and return useful results to the user, while also managing potential privacy concerns.

To address these shortcomings, the approach presented in this paper leverages a \emph{language-encoded} episodic memory agent that consists of three essential components that are demonstrated in Figure~\ref{fig:teaser}. First, we use language encoding to represent egocentric videos by harnessing the power of a large vision language model. Second, we store the language embeddings in a vector database for efficient retrieval. Third, we leverage a large language model to facilitate open-ended question answering in episodic memory tasks, using the vector database to enable the language model to access encoded memories. Using this approach, we focus on fine-tuning a large vision language model called Large Language and Vision Assistant (LLaVA)~\cite{liu2023visual}, specifically for egocentric data. 
For question answering (QA) in episodic memory, we use OpenAI GPT-4~\cite{gpt4} and integrate it with Chroma~\cite{chroma2023}, a vector database that enables long-term memory storage and retrieval.

Barman et al.~\cite{barmann2022did} introduced the QA-Ego4D dataset as a novel dataset designed for the memory augmentation task. It is offered as an extension of the Ego4D dataset~\cite{grauman2022ego4d}.
Each sample in the QA-Ego4D dataset consists of a video, a natural language question, and an answer.
The QA-Ego4D dataset is unique in its focus on the Episodic Memory Question Answering (EMQA) task, which introduces a constant-size memory constraint. 
This makes it particularly suitable for scenarios where the video content has long duration, or the system needs to be used in a life-long manner.
Barman et al.~\cite{barmann2022did} used multiple conventional vision-based machine learning models for this task, yet they demonstrated substantial limitations. Their performance barely surpassed a random guessing strategy, attaining BLEU scores between 3.4 and 5.8. In comparison, the method introduced in this paper achieves state-of-the-art results, outperforming the models by Barman et al.~\cite{barmann2022did} with a notable BLEU score of 8.3 for this task.

To assess the usability of our agent, we implemented it on the HoloLens 2 device, an AR headset, and allowed participants to ask open-ended questions to probe the agent's capabilities.
\changetext{
The results of this evaluation underscore the agent's practicality and acceptance among users.
}

Hence, our research paper presents three contributions as outlined below:
\begin{s_itemize}
    \item To the best of our knowledge, we present the first memory augmentation agent that integrates vision language encoding with episodic memory tasks, while also using a vector database for efficient storage and retrieval.
    \item We report the results of a lagre-scale quantitative evaluation of our agent using the Episodic Memory QA benchmark, specifically with the QA-Ego4D dataset. The results of these evaluations demonstrate the effectiveness of our agent in tackling episodic memory tasks.
    \item We report the results of a user study to explore the application potential of our agent. Our study showcases the benefits of our agent in various scenarios. By involving users and obtaining their feedback, we gain valuable insights into the practical implications and user satisfaction associated with our agent.
\end{s_itemize}

\section{Related Work}

\subsection{Augmented Reality: Augmenting User Perceptions}
\changetext{
Augmented reality (AR) is a technology that enriches the user's perception of the world by superimposing virtual content onto their physical environment~\cite{milgram1994taxonomy}. Typically, AR enhances user vision by overlaying digital images, videos, or 3D models onto real-world objects or scenes, creating an immersive and interactive experience~\cite{azuma1997survey}. However, AR encompasses more than just augmenting vision, as relying solely on visual augmentation would exclude individuals with disabilities, particularly those who are blind or have visual impairments~\cite{creed2024inclusive}. AR technology has the potential to be inclusive by extending augmentation beyond vision to other sensory modalities, such as auditory or tactile feedback~\cite{liu2018augmented}. For instance, AR applications could provide audio descriptions or tactile cues to convey information about the user's surroundings, enabling people with visual impairments to navigate and interact with the augmented environment~\cite{lo2021navigation,katz2012navig,troncoso2020aiguide}. By embracing multimodal augmentation, AR becomes accessible to a broader range of users, ensuring that everyone can benefit from its immersive and interactive experiences, regardless of their abilities or limitations.
}

\changetext{
Memory augmentation, for instance, is a notable application where AR can assist users in recalling information more effectively~\cite{rosello2016nevermind}. By providing contextual cues and audio feedback, AR can help users remember important details, tasks, or information associated with specific objects or locations. This broader scope of AR highlights its potential to enhance various facets of human perception and cognition, making it a versatile and impactful technology in numerous fields.
}

\subsection{Memory Augmentation through Lifelogging}
Lifelogging, present for over 30 years, has evolved with technology~\cite{harvey2016remembering,gurrin2013exploring}. Initially, it involved bulky equipment like helmets and battery packs~\cite{wolf2014lifelogging}, but has since progressed to wearable devices like glasses~\cite{harvey2016remembering}. A key development was Microsoft's SenseCam in 2006, a notable lifelogging device~\cite{SenseCam,doherty2012experiences}. Lifelogging now includes data from GPS, audio, heart rate, emails, calendar events, and social media. Significant progress has been made in memory augmentation through lifelogging~\cite{harvey2016remembering}. Le et al.~\cite{le2016impact} focused on video summaries for memory recall but didn't address data selection challenges. Davie et al.~\cite{Davies2015Security} highlighted privacy and security concerns in pervasive computing but lacked comprehensive solutions. Byrne et al.~\cite{byrne2010everyday} developed a method for content relevance in visual lifelogs but it was limited to everyday concepts. Our work uniquely implements a memory augmentation agent enabling open-ended episodic memory queries within a wearable headset.

\subsection{Video Content Analysis}

\label{sec:video_content}

Natural Language Video Localization (NLVL) and Video Question Answering (VideoQA) are distinct yet related tasks in video content analysis. NLVL focuses on finding a video segment matching a natural language query, requiring the model to understand both video and query context~\cite{krishna2017dense,gao2017tall,regneri2013grounding,grauman2022ego4d}. VideoQA, on the other hand, involves answering questions based on video content, demanding a deep understanding of the video and the ability to provide precise answers~\cite{lei2020tvqa, mun2020justask, sun2021video, miyanishi2021watch, le2020hierarchical}.

Episodic Memory Question Answering (EMQA) is a specific subtask of VideoQA, introduced by B{\"a}rmann et al.~\cite{barmann2022did}. It differs in its memory constraints, shifting from offline analysis (VideoQA) to an online algorithm and setting a maximum limit on memory usage for computation, thus suitable for long-term or life-long use.

This paper focuses on EMQA due to its advantages over NLVL and traditional VideoQA. While NLVL produces non-textual output and VideoQA has scalability issues, EMQA offers textual outputs and a constant-size memory constraint, enhancing efficiency for long videos.

\section{Agent Design}
Human memory involves encoding, storing, and retrieving information. Encoding is key for converting information into a format suitable for memory storage. Storage maintains this information until needed, and retrieval accesses and reinstates it into consciousness. Our agent mimics these biological memory processes~\cite{zlotnik2019memory}.

Initially, each video frame \( v \) is transformed into a language encoding \( l \) using the encoding function \( E \), denoted as 
$ l = E(v) $.
These language encodings are accumulated over time, forming a cumulative history \( L_{\text{history}} \).

An embedding model acts as a transformation function \( T \) to convert chunks $C$ of \( L_{\text{history}} \) into vector representation \( g \), expressed as 
$ g = T(C) $.
The vector \( g \) is stored in a vector database via a storage function \( S \), where 
$ S(g) \rightarrow \text{Database} $.

For retrieval, the agent uses the same transformation function \( T \) to convert a natural language query \( q \) into a query vector \( q_v \):
$ q_v = T(q) $.
The retrieval function \( R \) then fetches the most relevant language encodings \( c \) as context based on \( q_v \), formulated as 
$ R(q_v, \text{Database}) \rightarrow c $.
This context \(c \) and the query \( q \) are concatenated and processed by large language models to generate an answer based on the context.

\subsection{Encode}
We employ language as a means to encode egocentric visual perceptions. Specifically, our focus lies on adopting a frame-based approach rather than encoding clips using a sliding window. This decision stems from the fact that encoding clips using a sliding window can result in excessively long inference times, rendering it impractical for real-time usage.
Furthermore, the field of video captioning is still in its early stages, and even state-of-the-art models are unable to provide accurate and detailed encodings~\cite{xu2023mplug,damonlpsg2023videollama,xu2016msr,wang2022git}. Consequently, we opt to encode videos by their individual frames.

\begin{figure*}[t]
    \centering
    \includegraphics[width=0.5\textwidth]{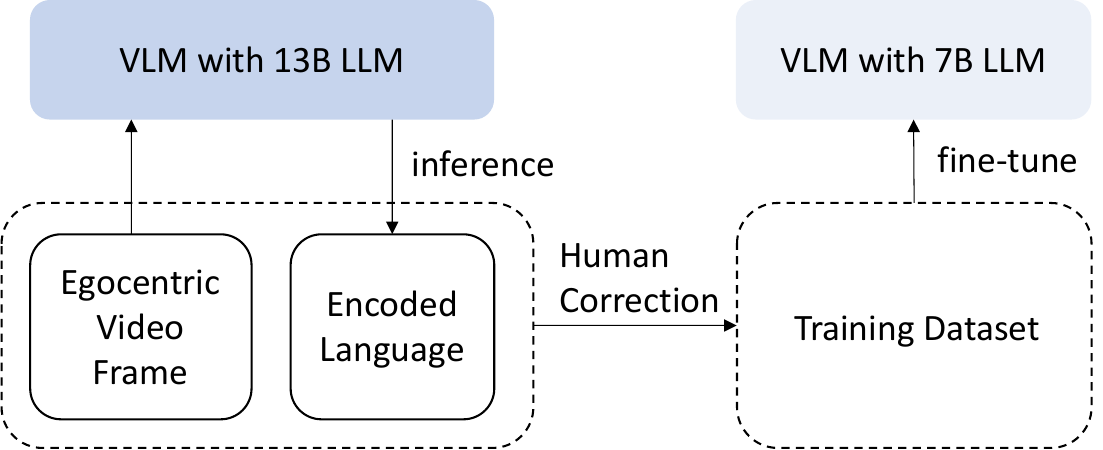}
    \caption{The Egocentric Vision-Language Model is developed through a process called fine-tuning. This process involves extracting knowledge from a large model and transferring it to smaller models, resulting in improved accuracy and faster inference times. The Egocentric Vision-Language Model combines the power of vision and language to effectively process and understand egocentric video data. 13B and 7B refer to large language models with 13 billion and 7 billion parameters.}
    \label{fig:model_distillation}
\end{figure*}
We present a novel model, Ego-LLaVA, for egocentric video encoding.
This model is fine-tuned from LLaVA (Large Language and Vision Assistant)~\cite{liu2023visual} on egocentric data, which captures first-person experiences in a 3D environment. This fine-tuning procedure leads to better performance in understanding first-person data which involves interpreting human-object interactions and complex social behaviors.

To tackle this issue, we curated our own egocentric video frame description dataset from Ego4D~\cite{grauman2022ego4d} and fine-tuned the LLaVA model to learn egocentric features.

The fine-tuning process is described below:
\begin{s_itemize}
    \item \textbf{Training Data}: 
    We begin by employing LLaVA using a descriptor prompt $P$, to generate detailed descriptions for a randomly sampled set of 3,000 images. Subsequently, we engaged three research assistants from our institution to correct the descriptions in scenarios where objects were inaccurately identified or significant objects within the frames were missed. This process results in a collection of 3,000 image/video frame-text/description pairs.
    It has been observed by Zhu et al.~\cite{zhu2023minigpt} that a set of 3,000 training pairs is adequate. Zhu et al.~\cite{zhu2023minigpt} successfully fine-tuned a visual-language model using only 3,500 image-text pairs, which yielded exceptional performance in tasks such as image question answering.

    The practice of training language models using responses generated by larger language models has become increasingly common due to the robustness of these models.
    Vicuna-13B~\cite{chiang2023vicuna} is an example of a model trained by fine-tuning the LLaMA-13B~\cite{damonlpsg2023videollama} base model with approximately 70,000 user-shared conversations gathered from ShareGPT~\cite{ShareGPT}, a website that collects conversational data from OpenAI ChatGPT. Similarly, MiniGPT-4~\cite{zhu2023minigpt} and LLaVA~\cite{liu2023visual} are trained using large language model-generated content, achieving state-of-the-art results and saving significant time on human labeling.
    
    \item \textbf{Fine-Tuning}: 
    In our experiment, as shown in Figure~\ref{fig:model_distillation}, we use Vicuna-13B~\cite{chiang2023vicuna} as the 13B language model and MPT-7B~\cite{MosaicML} as the 7B language model. MPT (MosaicML Pretrained Transformer) is optimized for efficient training and fast inference, utilizing FlashAttention~\cite{dao2022flashattention} and FasterTransformer~\cite{FasterTransformer} techniques.
    
    More specifically, Ego-LLaVA is fine-tuned on image-text pairs where the descriptor question $P$ prompts a description of the video frame $v$, and the ground truth prediction answer $l$ is the original detailed description. During training, the weights of both the visual encoder and LLM are kept constant, and the probability $p(l|v, P)$ of the target answers $l$ is maximized by only training the parameters of the linear projection layer between the visual encoder and the LLM.
    This process allows for the alignment of the video frame features $H_v$ with the pre-trained LLM word embedding.

\end{s_itemize}

\subsection{Store and Retrieve}
\label{sec:store}

Our approach involves segmenting language-encoded data into fixed-size chunks of 1024 tokens with a 256-token overlap to maintain semantic context, discarding any chunks under five characters. We attach relevant metadata to each chunk for enhanced search capabilities in vector databases. For capturing semantic meanings, we use OpenAI's \textit{text-embedding-ada-002} to create vector embeddings, which are then stored in the Chroma~\cite{chroma2023} vector database. This setup facilitates efficient management and retrieval of high-dimensional vector data.

The retrieval system operates by converting a user's input question into an embedding using OpenAI's \textit{text-embedding-ada-002} model. This embedding serves as a query to the Chroma~\cite{chroma2023} database, which searches for closely related vector-indexed data chunks. These chunks are then incorporated into a prompt along with the question for OpenAI GPT-4 to generate a response~\cite{stata2000term,langchain}.

\section{Study 1: Large-Scale Evaluation of the Memory Augmentation Agent}

To study the proposed memory augmentation agent's performance we carry out a large-scale quantitative evaluation using the public dataset QA-Ego4D. The evaluation focuses on the EMQA task, which is detailed in Section~\ref{sec:video_content}.
\begin{figure}[t]
    \centering
    \includegraphics[width=0.9\textwidth]{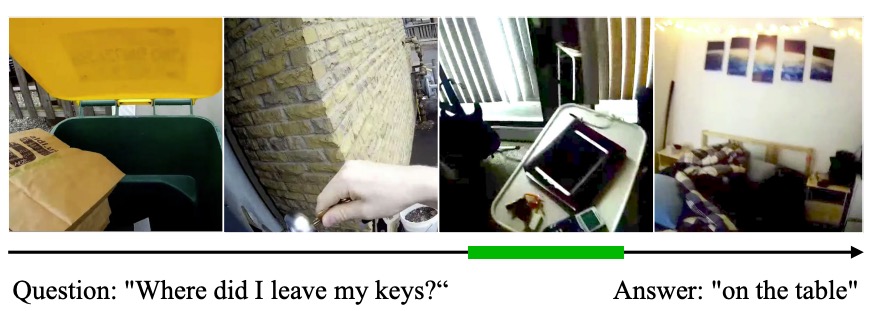}
    \caption{An example QA pair from the QA-Ego4D dataset adopted from~\cite{barmann2022did}.}
    \label{fig:example}
\end{figure}

\subsection{Dataset - QA-Ego4D}
The QA-Ego4D dataset, an extension of the Ego4D dataset's Natural Language Query (NLQ) subtask, features egocentric videos paired with natural language questions, answers, and annotations for answer-relevant video segments~\cite{barmann2022did, grauman2022ego4d}. 
Figure~\ref{fig:example} illustrates an example QA pair from the QA-Ego4D dataset~\cite{barmann2022did}.
Each video averages eight minutes in length. The dataset includes 19.2K queries from 227 hours of video across 34 scenarios from ten universities. Queries average 8.3 words, with response windows averaging 9.3 seconds, presenting a search challenge. The dataset omits ``When?'' questions due to undefined natural language answers.

It's divided into training, validation, and test sets, with 997 training videos, 162 for validation, and 166 for testing, comprising 10,746, 1,913, and 1,854 question-answer pairs for each set respectively. The test data uses half of the validation set's canonical videos, as Ego4D's test data is unpublished.
\changetextt{Explicit measures were implemented during the splitting process of the train and validation datasets to avoid any overlap, thereby preventing model overfitting~\cite{grauman2022ego4d}.}

\subsection{Baseline Models}

In our comparison, we include models from the QA-Ego4D paper~\cite{barmann2022did}: Differentiable Neural Computer (DNC)~\cite{graves2016hybrid}, Self-attentive-Associative-Memory-based Two-memory Model (STM)~\cite{le2020self}, Long-Term Comprehensive Transformer (LT-CT)~\cite{rae2019compressive}, and Rehearsal Memory (RM)~\cite{zhang2021learning}.

We also employ alterations to the encoding methods:
\begin{itemize}
    \item Language-Encoded QA (with Video-LLaMA~\cite{damonlpsg2023videollama}): We employ a sliding window approach with a width and stride of 6 seconds each to encode video clips into language. Video-LLaMA is a state-of-the-art video QA model which is suitable for video captioning.
    \item Language-Encoded QA (with LLaVA~\cite{liu2023visual}): We use the original LLaVA as the encoding method.
\end{itemize} 
We prompt both the above two models using the same prompt as for Ego-LLaVA.
These contrasting models provide a comprehensive comparison for the model proposed in this paper.

\subsection{Evaluation Metrics}
We report standard Natural Language Processing metrics for EMQA tasks, including BLEU-4~\cite{papineni2002bleu}, METEOR~\cite{banerjee2005meteor}, and ROUGE-L (f-score)~\cite{lin2004rouge}. These metrics all measure the similarity between a machine-generated sentence and a reference sentence. We report the values in percentage values. A higher value indicates better performance. 
\changetextt{
More specifically: BLEU-4~\cite{papineni2002bleu} measures the precision of n-grams (up to four words) in the generated text against the reference text, METEOR~\cite{banerjee2005meteor} evaluates based on precision, recall, and alignment, incorporating synonymy and stemming, and ROUGE-L (f-score)~\cite{lin2004rouge} assesses the longest common subsequence between generated and reference texts.
}





\subsection{Results}

\begin{table}[t]
\centering
\caption{EMQA results on the QA-Ego4D test set.}
\begin{tabular}{cccc}
\hline
\textbf{Model}                                                                          & \textbf{BLEU} & \textbf{METEOR} & \textbf{ROUGE} \\ \hline
DNC~\cite{graves2016hybrid}                                                                                & 3.4           & 17.9            & 27.0           \\
STM~\cite{le2020self}                                                                                & 5.8           & 17.6            & 26.2           \\
LT-CT~\cite{rae2019compressive}                                                                              & 5.3           & 18.5            & 27.5           \\
RM~\cite{zhang2021learning}                                                                                 & 4.5           & 17.7            & 26.6           \\ \hline
\begin{tabular}[c]{@{}c@{}}Language-Encoded QA\\ (with Video-LLaMA~\cite{damonlpsg2023videollama})\end{tabular}              & 5.8           & 19.3            & 30.7           \\
\begin{tabular}[c]{@{}c@{}}Language-Encoded QA\\ (with LLaVA~\cite{liu2023visual})\end{tabular}              & 7.4           & 36.1            & 50.7           \\
\textbf{\begin{tabular}[c]{@{}c@{}}Language-Encoded QA\\ (With Ego-LLaVA)\end{tabular}} & \textbf{8.3}  & \textbf{42.3}   & \textbf{54.7}  \\ \hline
\end{tabular}
\label{tab:emqa_results}
\end{table}

Table~\ref{tab:emqa_results} shows the outcomes of the various methods explained above applied to the QA-Ego4D test set. 
We observe that the language-encoded approach significantly outperforms conventional machine learning models. Additionally, we find that Ego-LLaVA surpasses the original LLaVA as an encoding method. This improvement can be attributed to the fine-tuning of LLaVA using egocentric data.
Despite the relatively small size of the fine-tuning dataset, which consists of only 3,000 image-text pairs, the model successfully learns to align image embedding features with text features. The performance in using Video LLaMA as the encoding model is poor because it suffers from severe hallucination issues.

\begin{table}[t]
\centering
\caption{The templates span a wide range of inquiries that individuals can make use of to enhance their memory, and retrieve information about various objects, locations, and individuals they encounter in their daily lives. We also show the average BLEU score for the proposed memory augmentation agent for each template. \changetextt{We adopted the same scale as prior work~\cite{barmann2022did}, which uses a percentage as given by SacreBLEU~\cite{post-2018-call}.};}
\begin{tabular}{cll}
\hline
\multicolumn{1}{l}{\textbf{Category}} & \textbf{Template}                                                                         & \textbf{BLEU} \\ \hline
\multirow{9}{*}{Objects}              & \begin{tabular}[c]{@{}l@{}}Where is object X before \\ / after event Y?\end{tabular}      & 8.7  \\
                                      & Where is object X?                                                                        & 8.9  \\
                                      & What did I put in X?                                                                      & 7.6  \\
                                      & \begin{tabular}[c]{@{}l@{}}How many X’s? \\ (quantity question)\end{tabular}              & 7.4  \\
                                      & What X did I Y?                                                                           & 8.3  \\
                                      & \begin{tabular}[c]{@{}l@{}}In what location \\ did I see object X?\end{tabular}           & 8.5  \\
                                      & What X is Y                                                                               & 8.0  \\
                                      & State of an object                                                                        & 7.8  \\
                                      & Where is my object X?                                                                     & 9.0  \\ \hline
Place                                 & Where did I put X?                                                                        & 8.2  \\ \hline
\multirow{3}{*}{People}               & \begin{tabular}[c]{@{}l@{}}Who did I interact \\ with when I did activity X?\end{tabular} & 8.1  \\
                                      & \begin{tabular}[c]{@{}l@{}}Who did I talk to \\ in location X?\end{tabular}               & 8.4  \\\hline
\end{tabular}
\label{tab:template}
\end{table}

Table~\ref{tab:template} shows the agent's performance on various question templates. We observe that simpler questions, such as ``Where is object X?'' or ``Where did I put X?'', generally yield better results due to their straightforward nature, demanding less complex reasoning from the agent. Conversely, questions involving more intricate reasoning or understanding of dynamic elements, such as ``Where is object X before / after event Y?'' or ``What X did I Y?'', may not perform as well. This is attributed to the current encoding method's limitations in capturing temporal correlations, which are crucial for comprehending dynamic activities. Quantity-based questions, such as ``How many X’s?'', pose a challenge due to the encoding model's resolution limitations, making accurate object counting difficult. 

Questions about the state of an object could also be challenging if the state involves fine details, or dynamic elements that change over time. Without the ability to apply attention to the data, the agent might not capture these dynamic subtleties, leading to a significant loss of crucial information. In essence, the agent's performance on different question templates largely depends on the question's complexity, the required level of detail, and the agent's ability to understand dynamic elements and temporal correlations. Future improvements in these areas could potentially enhance the agent's performance on more complex question templates.

\section{Study 2: Usability study for open-ended questions}

\begin{figure*}[t]
    \centering
    \includegraphics[width=\textwidth]{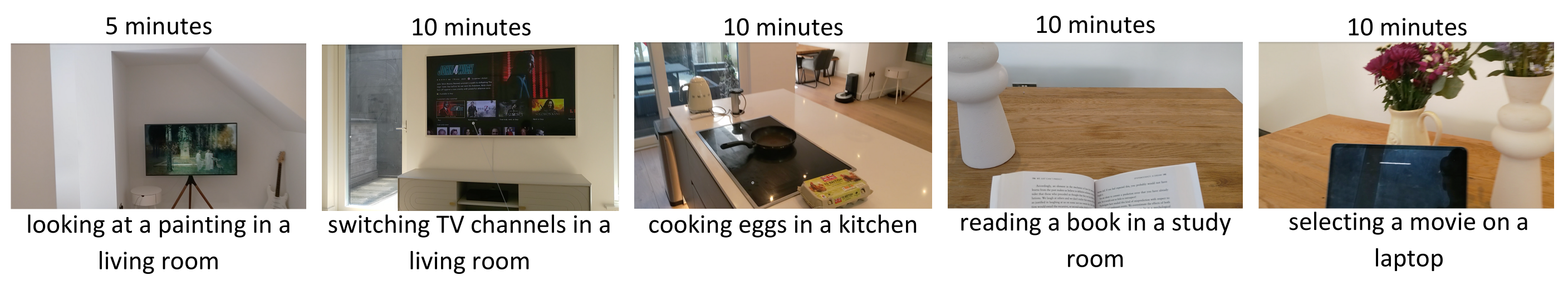}
    \caption{The settings of the different scenarios and the duration of each scenario.}
    \label{fig:study_setting}
\end{figure*}

Having established quantitative performance benefits of the memory augmentation agent in theory it is natural to ask whether the agent is usable in practice. 
\changetextt{
To this end, we carried out a user study  with two objectives. The first objective is to evaluate and contrast the performance of human participants with that of the memory augmentation agent in answering a set of episodic memory questions. The second objective is to explore the framework's capability in handling open-ended questions, which could potentially demand strong reasoning power and access to an external knowledge base. 
}
We did not incorporate another memory augmentation agent as a comparison because we are the first to propose such a memory augmentation framework and hence no such baseline exist. The conventional machine learning models have very limited usability as suggested by the results of the large-scale evaluation we described previously.

\subsection{Methodology}
\subsubsection{Participants}
We recruited a total of 12 participants using opportunity sampling to take part in the study (average age = 26.7, sd = 5.2; 7 males and 5 females).

A G*Power's analysis~\cite{faul2007gpower} based on a $t$-test suggested a sample size of 12 as being adequate for the study based on an effect size of 0.81 (calculated from the collected results), an error probability of 0.05, and a power of 0.8.
Among the 12 participants in the user study, four were students, five were employed, and three were self-employed. We envisage the intended users of this memory augmentation agent to be early adopters within the age range of 25 to 34, possessing AR/VR devices.

\subsubsection{Materials}
The study used a HoloLens 2 device which has an inbuilt front camera to stream egocentric videos. The encoding, storage, and retrieval tasks were performed by calling APIs hosted on our server.

\subsubsection{Protocol}
\begin{s_enumerate}
    \item \textbf{Study procedure}: The study consisted of two stages. During the first stage, participants were equipped with a HoloLens 2 device and were instructed to perform a series of tasks. These tasks were divided into five different scenarios: (1) looking at a painting in a living room; (2) switching TV channels in a living room; (3) cooking eggs in a kitchen; (4) reading a book in a study room; and (5) selecting a movie on a laptop. These tasks took place in an actual house equipped with a variety of furniture and items. \changetextt{Figure~\ref{fig:study_setting} illustrates the settings of the different scenarios and indicate the duration of each scenario.}
    Participants were encouraged to freely engage in the tasks to simulate a normal daily life experiences. Between five and seven days later, participants proceeded to the second stage of the study. This delay was used based on the concept proposed by Rivera-Lares et al.~\cite{RiveraLares2022}. They suggest that after a week, the amount of retained information may have diminished to a level referred to as the ``floor'', making it challenging to detect or observe any additional instances of forgetting. This can be represented by the forgetting curve, which hypothesizes a decline in memory retention over time in the absence of deliberate attempts to retain information.
    \item \textbf{Episodic memory questions to the participants and the agent}:
    \changetextt{
    During the second stage of the study the participants were presented with a set of questions modeled after Table~\ref{tab:template}, which were derived from the tasks they performed. Standard questions included ``Where did you place the TV remote?'', ``Name the list of movies you browsed on the laptop?'', ``What was the dominant color of the painting you observed?'', ``How many eggs did you cook?'', and ``Name the book you read?''. In addition to these task-related questions, there were other queries that were not directly linked to the tasks but remained relevant to the scenarios, including ``What color was the guitar beside the painting?'', ``What was the person you interacted with (study facilitator) wearing?'', ``What is the color of the kettle beside the pan'' and ``How many vases did you see on the dining table?''. An example of a description type question for the third category of queries is the following: ``Describe the painting in detail''. We asked each participant these ten questions. Same questions were used to ask the memory augmentation agent to generate responses. 
    }
    \item \textbf{Open-ended questions to the agent}: 
    Subsequently, participants were encouraged to ask the memory augmentation agent five open-ended questions through an interactive conversational interface. Examples of questions were ``What movie would you recommend for next time?'', ``Based on what you know, do you think I eat healthily, and if not, what suggestions do you have for my diet?'', and ``What are the steps to better cook an egg?''. 
    \item \textbf{Rate the agent's responses}:
    \changetextt{
    Thereafter the participants were asked to rate the responses generated by the agent as well as their own answers using a scale ranging from 1 (very bad) to 5 (very good). Note that the order of queries and the order of which responses to score were randomized. Participants were also requested to score the agent's responses to open-ended questions using the same scale.}
    \item \textbf{Post-study Likert scale questions}: 
    We then used post-study questionnaire to gather feedback from the participants regarding their subjective opinions on the overall experience with the agent. The participants were asked to respond to the following Likert scale questions: (1) The memory augmentation capability is valuable; (2) The information provided by the memory augmentation agent is accurate; (3) The response to my open-ended questions by the memory augmentation agent is creative; (4) I am willing to wear an always-on camera for language-encoded memory augmentation; and (5) I am willing for others in my close vicinity to wear an always-on camera for language-encoded memory augmentation. The participant responses are distributed across five categories: Strongly Disagree, Disagree, Neutral, Agree, and Strongly Agree, for different aspects of the memory augmentation agent, including its capability, accuracy, creativity, and willingness to use. 
    \item \textbf{Open-ended survey}: Finally, we asked four open-ended questions: (1) Under what circumstances would you use this memory augmentation feature?; (2) Do you have any concerns regarding the memory augmentation capability?; (3) What improvements would you suggest for the memory augmentation agent? and (4) Do you have any other feedback or suggestions regarding the memory augmentation feature? 
\end{s_enumerate}

\subsubsection{Implementation}
The process begun with the uploading of the video to the server. To optimize the speed of the encoding process, we extracted four frames per second from the video. To expedite this process and approach near real-time encoding, we used multi-process threading. This technique allows multiple encoding tasks to be executed simultaneously, significantly reducing the overall time required.
Once the frames were encoded, they were stored in a vector database. 
We used LangChain~\cite{langchain} and Chroma~\cite{chroma2023} as the implementation framework for vector data storage and retrieval, and OpenAI GPT-4 as the large language model to implement the conversational AI assistant that performs question-answering for memory augmentation.

\begin{figure*}[t]
    \centering
    \includegraphics[width=0.8\textwidth]{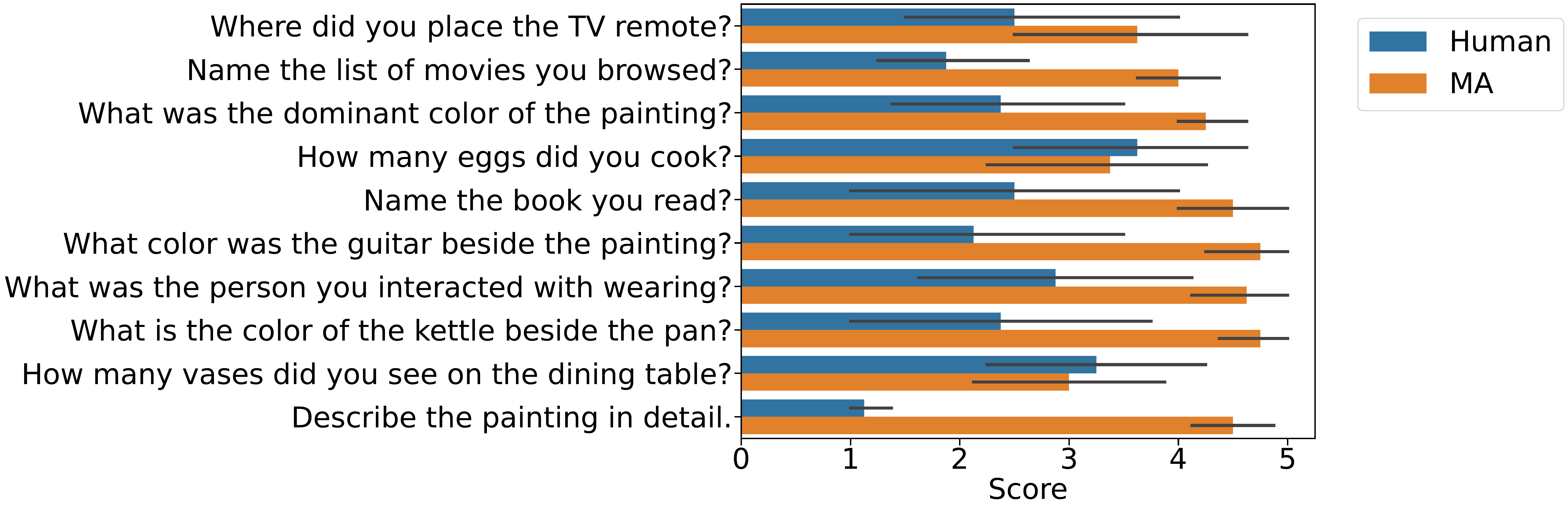}
    \caption{Comparative analysis of scores for the memory augmentation agent and Human responses across various questions. Each question has multiple pairs of AI and human scores represented by the bars. The $x$-axis enumerates different questions, while the $y$-axis shows the scores ranging from 1 to 5. The bars are color-coded, with one color representing AI and another representing human scores. The legend on the top-right corner outside the plot area distinguishes between AI and human bars.}
    \label{fig:response}
\end{figure*}

\begin{figure}[t]
    \centering
    \includegraphics[width=\textwidth]{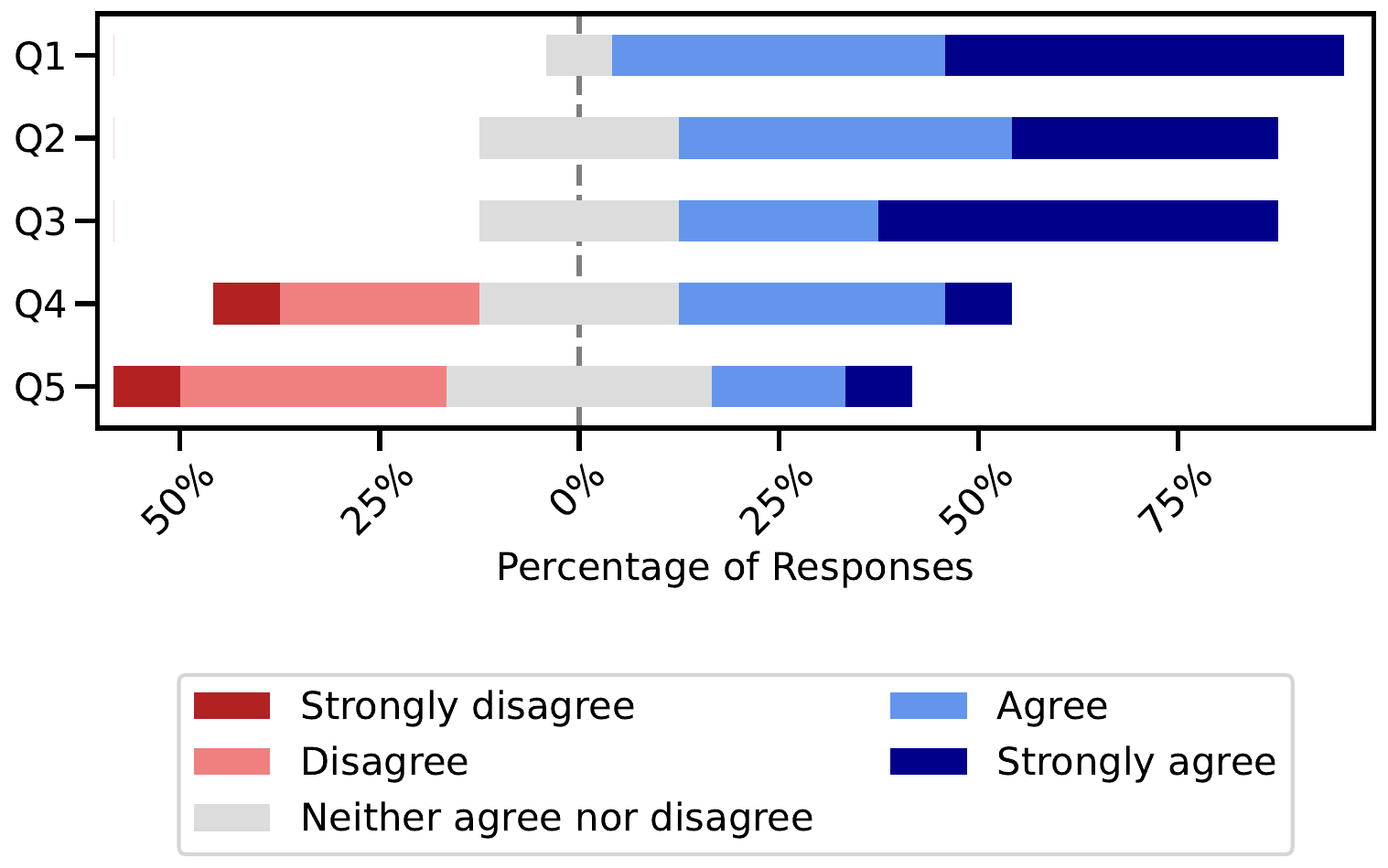}
    \caption{The five-point Likert responses to the post-study questionnaire. Q1. The memory augmentation capability is valuable; Q2. The information provided by the memory augmentation agent is accurate; Q3. The response to my open-ended question by the memory augmentation agent is creative; Q4. I am willing to wear an always-on camera for memory augmentation through language encoding; Q5. I am willing for others in my close vicinity to wear an always-on camera for memory augmentation through language encoding.}
    \label{fig:likert}
\end{figure}

\subsection{Results}
\label{sec:user_study_results}

    \subsubsection{Episodic Memory Questions}
    \changetextt{
    Figure~\ref{fig:response} illustrates a comparison of scores for the memory augmentation agent's response and human responses across various episodic memory questions. 
    To clarify how a rating score corresponds with a response, we relate findings to the scored responses. Consider an example question: ``What was the person you interacted with wearing?'' An answer of ``grey t-shirt'' would receive a score of 5/5, while a response of ``grey'' would score 4/5, and a completely unrelated answer, such as ``red'' or ``I don't know'' would receive a score of 1/5. For color and quantity questions, the closer the answer is to the actual response, the higher the score. For instance, ``blue'' would be rated around 4/5 in response to the color ``green'', and a response of ``4'' would get a score around 4/5 for a quantity of ``5''.
    The memory augmentation agent outperforms humans in general, as evidenced by the performance shown in Figure~\ref{fig:response}, particularly for  standard episodic memory questions. 
    It excels in tasks that require detailed descriptions, as it demonstrates the capability to provide thorough descriptions of paintings. The ego-LLaVA decoding method surpasses human memory in capturing finer details. 
    }
    
    \changetextt{
    However, the language encoding method exhibits a weakness in tasks that involve counting, such as determining the number of eggs or vases. Moreover, in tasks involving dynamic elements, such as the question ``Where did you place the TV remote?'', while the ego-LLaVA accurately identifies the TV remote, it encounters a failure mode in distinguishing between ``pick up'' and ``place''. Sometimes, it provides an answer based on the former action, and when participants do not place the TV remote in the same location, the agent may provide an incorrect response.
    We also observed that human memory is highly proficient in specific tasks but the participants tended to struggle with remembering things they did not pay attention to, such as the color of a kettle, or the presence of a guitar, when unrelated to their current tasks. Additionally, recalling a list of five movie names poses a challenge for human memory, resulting in subpar performance.
    }
    
    \changetextt{
    In summary, the memory augmentation agent proves useful in several scenarios, including: (1) memory-intensive tasks; (2) tasks requiring detailed descriptions; and (3) situations where people do not focus on, or pay attention to, certain aspects.
    The memory augmentation agent yielded a higher mean response with 4.1/5 compared to a human response of 2.5/5, indicating its superiority over human performance. A Friedman's test revealed a statistically significant difference between the memory augmentation and human responses ($\chi^2 = 37.928$, df = 1, p = 0.0009), providing evidence to reject the null hypothesis of no difference. Further, the memory augmentation agent resulted in a smaller standard deviation (1.1/5, compared with 1.7/5 for humans), suggesting increased stability in  responses compared to the less reliable human memory. Overall, the memory augmentation agent performs better and exhibits greater consistency compared to human responses.
    }
    \subsubsection{Open-Ended Questions}
    Participants rated the responses provided by the memory augmentation agent in response to open-ended questions at an average score of 3.97/5, and a standard deviation of 0.604. This suggests the additional value of using a large language model like OpenAI GPT-4 as a conversational interface with access to an external knowledge base, enhancing its context awareness regarding memories. Some frequent questions are to recommend a movie or a book. Then the agent responded with ``Certainly! Based on the context provided, I suggest you watch The Godfather. It's a classic and highly acclaimed film that you might enjoy.'', or ``I would recommend checking out The Dark Knight as it seems to be a popular choice in the given scenarios. It is a movie and comic book series that you might find interesting.'' These responses are usually rated as 5/5. There are also questions related to instructions, such as ``How can I cook my egg better?''. The memory augmentation agent responds by listing very detailed steps for cooking the eggs while also being aware of the quantity of the eggs the user cooks. Such responses are often rated as 4/5 to 5/5.

    \subsubsection{Post-Study Likert Scale Questions}
    Figure~\ref{fig:likert} shows the distribution of the participant responses for the memory augmentation agent. \changetext{Participants value the memory augmentation agent, indicating a strong recognition of its utility in enhancing memory recall effectively. The accuracy of the information provided by the agent also received positive feedback, affirming its reliability. Responses about the agent's creativity, though mixed, still highlight a level of engagement that could be built upon. Some participants showed a willingness to wear an always-on camera, suggesting an openness to integrating this technology into daily life despite potential privacy concerns. This acceptance is mirrored in their comfort with others using the technology, pointing towards its feasibility in social or communal settings.}

    \subsubsection{Open-Ended Survey}
    From the open-ended survey, we observed that participants emphasized the importance and value of having a memory augmentation agent, highlighting various applicable scenarios such as biomedical experiments, conferences, attending lectures/meetings, and exploring new places \textit{(P1, P3, P4, P6, P7, P10, P11, P12)}.
    However, participants expressed concerns about the agent being socially awkward to wear \textit{(P1, P2, P6)}.
    Ethical concerns were also raised, such as the potential degradation of people's memorization capabilities if they rely solely on the agent \textit{(P4, P10)}.
    Additionally, some participants highlight privacy issues concerning individuals donning it and recording their actions. 
    For example, \textit{P3} noted that ``the agent's powerful and accurate capabilities could pose safety risks if breached'', while \textit{P2} expressed worries about ``discomfort with others wearing the system and recording their activities''.

    However, as the understanding of the agent's function through language encoding grew, the majority of concerns  diminished \textit{(P1, P3, P5, P6, P7, P8, P12)}, although a few participants still felt uneasy \textit{(P1, P2, P9, P11)}.
    Participants proposed specific improvements to address these issues, including incorporating indicators to make people aware of the agent's operation and limiting its use to specific scenarios such as teaching and conferences rather than everyday life \textit{(P1, P3, P4, P12)}. Additionally, participants suggested making the always-on camera device as lightweight and inconspicuous as possible to minimize social awkwardness and increase social acceptance \textit{(P2, P5, P6, P8, P9, P11)}.


\section{Discussion}
In this paper we have demonstrated a novel memory augmentation agent and demonstrated its performance. Besides being lightweight, as it is reliant on language-encoding as opposed to vision-based, we discuss three additional advantages of this approach: performance, privacy, and device agnosticism.

\subsection{Ego-LLaVA Paired with Language Encoding Achieved State-of-the-Art in QA-Ego4D}

The evaluation of the system using the QA-Ego4D dataset demonstrated superior performance, surpassing traditional machine learning models with a substantial BLEU score of 8.3. On the other hand, the baseline models exhibit subpar performance on the EMQA task due to primarily three factors.
First, memory constraints: the EMQA task imposes a constant-size memory constraint, requiring models to compress all potentially relevant information into a fixed-size representation. Baseline models not designed to handle this constraint may struggle to effectively manage and utilize the limited memory space.
Second, model specialization: certain baseline models, such as STM, use a simplistic implementation that may not be sufficient for the EMQA task. For example, STM's reliance on a single hidden vector in its implementation could account for its low performance.
Third, relevance selection: the EMQA task necessitates the model to appropriately select relevant information due to the memory constraints. This could be a challenge for some models, leading to poor performance.

\subsection{Language Encoding is Lightweight}
In this paper, the Language Encoding Approach and the Vision-based Approach are compared. Language Encoding stores textual data from video, requiring around 0.517 TB/year uncompressed, reduced to 0.246-0.345 TB with compression, while the Vision-based Approach needs 5.74 TB/year for low bitrate 720p video.

\subsection{Language Encoding is Device Agnostic}
\changetext{For question and answering, the language encoding approach introduced in this paper gives rise device agnosticism due to its design.
This contrasts with vision-based QA models which may exhibit diminished accuracy, or necessitate fine-tuning, when transitioning across different devices. 
Moreover, the device-agnostic nature is carried through in our language encoding model. 
The novel egocentric vision language model we introduced in this paper is cultivated using a diverse array of devices including GoPro, Vuzix Blade, Pupil Labs, ZShades, ORDRO EP6, iVue Rincon 1080, and Weeview, to capture egocentric videos. This breadth of training sources fortifies our framework's compatibility with any device capable of delivering egocentric video streaming. While we use the HoloLens 2 as the AR headset, its usage is solely as a conduit for streaming egocentric videos, further illustrating the adaptability of the model.}

\section{Limitations and Future Work}

We introduce an encoding method that operates on an individual frame basis. However, this method struggles to capture temporal correlations, which are essential for understanding dynamic features like activities. \changetextt{Figure~\ref{fig:response} shows the lower performance on the two questions related to dynamic elements, including ``place'' and ``cook''. Such movements or scene changes, are better understood when temporal correlations between frames are considered.} Without this, the encoding method may miss these dynamic subtleties, leading to a significant loss of crucial information. Activities usually occur over a series of frames. Ignoring temporal correlations can make it challenging to fully understand these activities. For example, the action of a person picking up an object involves a sequence of movements across several frames. Despite these shortcomings, this encoding method excels in capturing static features, as each frame is encoded separately.

\changetextt{Another limitation is that while human memory augmentation has been shown to augment users' memory in a controlled lab setting, as suggested by the results in Study 2, we acknowledge that this has not been tested in real-world scenarios. Additionally, a longitudinal study is lacking to assess its long-term effectiveness.}

Looking forward, we suggest research could focus on two main areas. First, to develop a high-resolution, accurate egocentric vision-language model. This model will integrate vision and language processing from a first-person perspective, similar to human perception of their surroundings. The high-resolution aspect of this model will allow it to capture intricate details in visual data, potentially resulting in significant performance improvements.
Second, to create a vision-language model that can process dynamic video data. This model will provide a more holistic view of the environment by capturing temporal changes and movements over time. This approach will yield richer and more contextual information, enhancing the model's understanding of the visual scene. However, it is important to note that processing video data is more complex and computationally demanding than handling static images, presenting its own unique challenges.

\section{Conclusion}
We have presented, memory augmentation agent, a novel system that combines language encoding with episodic memory tasks. The agent incorporates several essential elements made possible using natural vision language encoders, including other features, such as long-term memory integration, a lightweight implementation using hosted APIs, and real-time inference capabilities. This integration of language encoding and episodic memory represents a state-of-the-art approach in the field, enabling enhanced performance and efficiency. Moreover, while privacy is a significant concern in lifelogging and memory augmentation agents, the use of language encoding presents a viable solution forward. By transforming video data into language data and carefully managing the encoding process, privacy can be effectively safeguarded while still providing a useful tool for memory augmentation. 
The agent was evaluated to verify that it is effective and efficient. In comparison to traditional machine learning models, the new model demonstrated superior performance on the QA-Ego4D dataset. 
\changetextt{Further, a user study revealed that the agent yielded statistically significant better performance when handling episodic memory tasks. 
Taken together, the results demonstrate the memory augmentation agent to be viable in practical applications and another example of how AI can work in tandem with AR to create new ways of supporting users in their daily lives.}


\bibliographystyle{abbrv-doi}

\bibliography{template}
\end{document}